\pdfoutput=1

\documentclass[11pt]{article}

\usepackage{EMNLP2023}

\usepackage{times}
\usepackage{latexsym}
\usepackage{soul}
\usepackage{amsmath}
\usepackage{amssymb}
\usepackage{booktabs}
\usepackage{pdfpages}

\usepackage[T1]{fontenc}
\usepackage{hyperref}

\usepackage[utf8]{inputenc}

\usepackage{microtype}

\usepackage{inconsolata}

\usepackage{lipsum}
\usepackage{CJKutf8}
\newcommand{\x}{\mathbf{x}}

\usepackage{graphicx}

\newcommand{\cjktext}[1]{\begin{CJK}{UTF8}{gbsn}#1\end{CJK}}

\newcommand\hsfixed[1]{{#1}}
\newcommand\jhgreat[1]{{#1}}  
\newcommand\hsfixedsecond[1]{{#1}}
\newcommand\hsr[1]{{#1}}

\title{Character-level Chinese Backpack Language Models}

\author{Hao Sun \\
  Stanford University\\
  \texttt{haosun@stanford.edu} \\\And
  John Hewitt \\
  Stanford University \\
  \texttt{johnhew@cs.stanford.edu} \\}

\begin{document}
\begin{CJK}{UTF8}{gbsn}
\maketitle

\begin{abstract}
\jhgreat{
The Backpack is a Transformer alternative shown to improve interpretability in English language modeling by decomposing predictions into a weighted sum of token sense components.
However, Backpacks' reliance on token-defined meaning raises questions as to their potential for languages other than English, a language for which subword tokenization provides a reasonable approximation for lexical items.
In this work, we train, evaluate, interpret, and control Backpack language models in character-tokenized Chinese, in which words are often composed of many characters.
We find that our (134M parameter) Chinese Backpack language model performs comparably to a (104M parameter) Transformer, and learns rich character-level meanings that log-additively compose to form word meanings.
In SimLex-style lexical semantic evaluations, simple averages of Backpack character senses outperform input embeddings from a Transformer.
We find that complex multi-character meanings are often formed by using the same per-character sense weights consistently across context.
Exploring interpretability-through control, we show that we can localize a source of gender bias in our Backpacks to specific character senses and intervene to reduce the bias.
}

\end{abstract}

\section{Introduction}

\begin{figure}[t]
    \includegraphics[scale=.8]{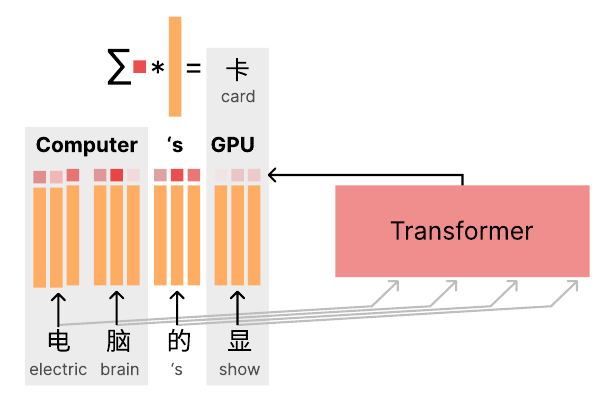}
    \caption{The general structure of the character-level Chinese Backpack Language Model. \hsfixedsecond{The next character is predicted by the weight sum of the senses of characters in the previous context. The sense vector of "显" (show) provides information for word composition, while the senses of "电" and "脑" (computer) provide semantic information through linear combination.} }
    \label{fig:gc}
\end{figure}

Language modeling is a crucial task in natural language processing, where the goal is to compute the probability of the next word in a sequence given the preceding words. Recently, large language models based on the Transformer architecture~\cite{vaswani2017attention} have achieved remarkable success in various NLP applications, including text generation~\cite{GPT2, brown2020language, gpt-j}, machine translation~\cite{bawden-etal-2019-university, lewis2019bart}, and question-answering~\cite{miller2017parlai, DBLP:journals/corr/abs-2004-04906, DBLP:journals/corr/abs-2101-00438}. 
However, Transformers are notoriously hard to interpret and control. Their non-linear contextualization functions imply that intervening on their internal activations can have unpredictable consequences.

The recently proposed Backpack architecture~\cite{backpack} tackles the interpretability problem by decomposing its predictions as a sum of non-contextual vectors, which then provide an interface for interpretability. Intuitively, it combines the expressivity of Transformers with some of the interpretability and control benefits of log-linear models.
It was shown to have similar \jhgreat{language} modeling capacity to Transformers on English, and performed comparably on perplexity and LAMBADA~\cite{paperno-EtAl:2016:P16-1} tests, at a tax of 1.4x more parameters. 

\hsr{The effectiveness of the Backpack architecture in languages with different morphological structures than English remains uncertain due to challenges in interpreting and controlling individual tokens without stable explicit semantics.
In Chinese, most vocabulary consists of compound words with multiple characters. However, these characters often have implicit meanings~\cite{packard2011new, 10.3389/fpsyg.2018.00258}, making it challenging to infer the meaning of these words based solely on the individual meanings of their constituent characters. Additionally, some characters represent the pronunciation of foreign words and lack semantic associations, which requires characters to learn more complex semantic connections within limited sense vectors.} 
The English-based Backpack model is trained on \jhgreat{often} complete words with \jhgreat{more} explicit meanings, \jhgreat{making it uncertain whether Backpacks will perform well in character-level Chinese.}

In this paper, we trained the first non-English (and first character-based language) Backpack language model and evaluate its performance and learned lexical semantics on character level.\footnote{\hsr{Our code, weights, and demos are available at \href{https://github.com/SwordElucidator/nanoBackpackLM}{https://github.com/SwordElucidator/nanoBackpackLM}}}  
We trained several Backpack and \hsr{Transformer} baseline models and evaluated them on perplexity and word prediction accuracy tasks. Our experiments show that our pretrained 134M Backpack Language Model with 16 sense vectors, which uses character-based tokenization, performs comparably to a 104M \hsr{Transformer} model. 

\jhgreat{To understand the Backpack's success, we first study how it composes word meaning from non-contextual token senses.}
\jhgreat{We hypothesize word meaning is formed because tokens of a multi-character word receive similar weighting in the Backpack's sum across all contexts the word appears in.} We find that indeed the proportion of these composed characters on each sense vector changes by no more than 20\% in over 90\% of cases. 
Moreover, we achieve better word representation 
\hsr{under three Chinese corpora} by simply averaging the sense vectors of composed characters compared to the character embeddings of the pretrained \hsr{Transformer} model. Additionally, we propose and evaluate character-level interventions to mitigate gender bias and control \jhgreat{how word meaning is composed from character meaning}
, which demonstrate promising results for generating controllable text in character-based Chinese Backpack models. \hsfixed{ These experiments 
\hsr{show} that our Chinese Backpack Model learns the implicit semantics of characters, making it possible to control the emphasis or weakening of certain characteristics of a word during generation tasks.}

\section{Related Work}
\subsection{Word Representation with Deep Learning}
Numerous word embedding techniques have been proposed in the early stages of natural language processing with deep learning, including Word2Vec~\cite{word2vec} and GLoVe~\cite{pennington2014glove}, which represent words as 
vectors.
Word2Vec 
learns word embeddings by predicting the probability of a word's occurrence given its context words or predicting the context words given a central word.
\citet{backpack} showed that the Backpack is a generalization of Word2Vec. 
While these methods produce high-quality word representations that capture the semantic and syntactic relationships between words and have enabled rich interpetability studies as well as bias auditing ~\cite{semantic,spine,bias}, they 
\hsr{are not suited to} language modeling tasks due to a lack of expressivity.

Subsequently, modern language models with 
the Transformer architecture~\cite{vaswani2017attention} 
build contextualized word embeddings \hsfixedsecond{that are useful for modeling language in a variety of settings.} 
\jhgreat{However, as noted by \citet{backpack}, these models' monolithic, non-linear processing of token sequences eschew any meaningful word-level semantics, so word-level interpretability has no direct connection to model behavior. Separately, interpreting contextual representations is difficult because each context maps arbitrarily to different representations}, making it difficult for word embeddings to directly represent non-contextual semantic information and challenging to achieve predictable intervention across all contexts.

\subsection{Language Modeling with Deep Learning}
\hsr{Language modeling} is a fundamental task in \hsr{natural language processing}
, involving computing the probability of the next word in a sequence given the previous words. Early 
\hsr{neural} approaches to language modeling \hsr{used feed-forward networks \cite{NIPS2000_728f206c}}, various Recurrent Neural Networks (RNNs)
\hsr{~\cite{elman1990finding,rnn}} and attention mechanisms~\cite{bahdanau2014neural}. More recently, modern language models have adopted the Transformer architecture~\cite{vaswani2017attention}, with the GPT series~\cite{radford2018improving, GPT2, brown2020language} by OpenAI achieving notable success in generating high-quality and coherent text. This success has led to applications in various areas, such as story generation~\cite{xu2020megatroncntrl,chen2021graphplan} and chatbots~\cite{lin2020caire,roller2020recipes,shuster2022blenderbot}. However, as previously discussed, interpreting word embeddings in Transformer-based language models poses a challenge.

\subsection{The Backpack Architecture}
\citet{backpack} introduced the Backpack, a neural architecture which achieves high performance on contextualization and non-contextual word representations. This approach represents each word in a sequence as a linear combination of sense vectors, with weights computed by an expressive network such as the Transformer. \jhgreat{(We'll review the Backpack in detail in Section~\ref{sec:approach}.)} The linearity of the contributions of sense vectors to predictions encourages the sense vectors to specialize and encode rich notions of word meaning during pretraining. Furthermore, the authors conducted experiments on sense vectors, demonstrating their potential for predictable control across all contexts. We reproduced and pretrained it on character-based Chinese language, demonstrating the Backpack model's potential for application to languages of this type.

\hsfixedsecond{\subsection{Chinese Tokenization and Embeddings}
One common approach for tokenization in Chinese involves sub-word tokenization methods, such as WordPiece~\cite{6289079}, byte pair encoding~\cite{sennrich-etal-2016-neural}, and unigram language model segmentation~\cite{kudo-2018-subword}, which were adopted by recent Chinese Pretrained Language Models such as CPM~\cite{zhang2020cpm}. Furthermore, \citet{si2023subcharacter} proposed Sub-Character Tokenization, which encodes each Chinese character into a sequence of phonetic or stroke symbols, and then utilizes a sub-word tokenization method to construct the vocabulary. In our research, to understand the performance of character-level sense vectors, we used single Chinese character tokenization method \hsr{proven to be 
effective by ~\citet{li-etal-2019-word-segmentation} and }utilized by \hsr{Chinese GPT2~\cite{GPT2-Chinese}} and MacBERT~\cite{Cui_2021, cui-etal-2019-span}.

\jhgreat{Various studies have} explored embeddings at the word~\cite{rumelhart1986learning, NIPS2000_728f206c, NIPS2008_1e056d2b}, phrase~\cite{Socher2010, Zhang2014BilinguallyconstrainedPE, Yu2015}, sentence~\cite{le2014distributed, socher-etal-2013-recursive, kalchbrenner2014convolutional}, and document~\cite{srivastava2013modeling, le2014distributed, hermann-blunsom-2014-multilingual} levels for representing knowledge and semantics. In the case of Chinese, character-level embeddings~\cite{chen2015joint, li-etal-2015-component} have also been investigated in relation to compounded word embeddings~\cite{xu-etal-2016-improve}. We investigated on character embeddings and conducted two methods for representing compounded words using the contextualization weights learned during pretraining.
}

\section{Approach} \label{sec:approach}
\subsection{Backpack language model}
\jhgreat{
Drawing directly from \citet{backpack}, a Backpack language model is a probabilistic model
\begin{align}
p(\mathbf{x}_i \mid \mathbf{x}_{<i}) = \text{softmax}(E^\top \mathbf{o}_{i-1}),
\end{align}
where $\mathbf{x}_{1:i}$ is a sequence of elements from finite vocabulary $\mathcal{V}$, $E\in\mathbb{R}^{d\times |\mathcal{V}|}$,  and $\mathbf{o}_{i-1}$ is a \textit{Backpack representation} of $\mathbf{x}_{<i}$.
In turn, a Backpack representation is constructed in two pieces:
\paragraph{Sense vectors.}
For each word in the vocabulary $\mathcal{V}$, a backpack learns $k$ \textit{sense vectors}, each like a specialized word2vec vector. 
We write the sense vectors for $\mathbf{x} \in \mathcal{V}$ as $\{C(\mathbf{x})_\ell\}_{\ell=1}^k$.
When presented with a sequence $\x_{1:i}$, the Backpack constructs its sense vectors for the words in the sequence:
\begin{align}
C(\x_1), \dots, C(\x_i).
\end{align}

\paragraph{Weighted sum.} 
The Backpack representation $\mathbf{o}_{i}$ is just a weighted sum of the sense vectors of the sequence:
\begin{align}
\mathbf{o}_i = \sum_{j=1}^{n} \sum_{\ell=1}^{k} \alpha_{\ell ij} C(\mathbf{x}_j)_\ell,
\end{align}
where $\alpha_{\ell ij}$ is defined by a contextualization function $\alpha = A(\mathbf{x}_{1:n})$, and $A: \mathcal{V}^n \rightarrow \mathbb{R}^{k \times n \times n}$, and all $\alpha_{\ell ij}\geq 0$. 

\subsection{A note on Backpack token semantics}
Intuitively, the contribution of each sense $C(\x)\ell$ to any prediction is \textit{independent of context}.
We find it instructive to write out what this means for token-level semantics.
The score $(E^\top\mathbf{o}_i)_{\mathbf{w}}$ of a word $\mathbf{w}\in\mathcal{V}$ in context $\x_{<i}$ is the unnormalized log-probability of that word.
Because of linearity, we have:
\begin{align}
E^\top \mathbf{o}_{i} = \sum_{j=1}^{n} \sum_{\ell=1}^{k} \alpha_{\ell ij} E^\top C(\mathbf{x}_j)_\ell,
\end{align}
The contribution of a sense $C(\x_{j'})_\ell$ to that word's score is thus
\begin{align}
\alpha_{\ell ij'} E^\top C(\x_{j'})_\ell \in \mathbb{R}^{|\mathcal{V}|}.
\end{align}
Because all $\alpha$ are non-negative, the \textit{meaning} or use of a sense is simply its set of scores over the vocabulary $E^\top C(\x_{j'})_\ell$, which depends only on the word (not the context); only the \textit{importance} of that meaning is determined by context.
As such, visualizations of the ``highest-scoring words'' for a sense---as we provide in future sections---have a particularly transparent connection to model behavior.

\subsection{Parameterizing Backpack Language Models}
}
%


The sense function is parameterized $C(x) = \text{FF}(Ex)$ where FF: $\mathbb{R}^d \rightarrow \mathbb{R}^{d \times k}$ is a a feed-forward network, and contextualization weights $A(\textbf{x}_{1:n}) = \alpha$ where 
\begin{align}
\alpha_\ell = \text{softmax}(\textbf{h}_{1:n}^\top K^{(\ell)\top} Q^{(\ell)}\textbf{h}_{1:n})
\end{align} 
for each predictive sense $\ell$ with matrices $K^{(\ell)}, Q^{(\ell)} \in \mathbb{R}^{d \times d / k}$ and $\textbf{h}_{1:n}$ calculated by a Transformer~\cite{vaswani2017attention} with autoregressive masking, i.e. 
\begin{align}
\textbf{h}_{1:n}=\text{Transformer}(E\textbf{x}_{1:n}) 
\end{align} 

\hsr{We} introduced a series of minor adjustments to the 
\hsr{implementation details of the} original backpack language model with the objective of enhancing training stability and facilitating a more comprehensive comparison between our model and the GPT model as discussed in Appendix~\ref{sec:modeling}.


\subsection{Baselines}
We employed a GPT2
\hsr{-like Transformer} model ~\cite{GPT2} as a baseline, pretrained using the same datasets, hyperparameters, and random seed as our Backpack model. The \hsr{Transformer} and Backpack models have equal contextual parameters in the Transformer structure, whereas the Backpack model contains additional non-contextual parameters for the sense vectors.
The \hsr{Transformer} and Backpack models share the same tokenizer and have an identical embedding size, as well as the same number of layers and heads for contextualization.

\section{Experiment Training Backpack LMs}

\hsfixed{To compare the performance of our models against the baseline models in general language modeling evaluations, We first pretrained our 134M "Backpack-small" and 27M "Backpack-micro" Chinese Backpack language models and the baseline 104M "GPT2-small" and 18M "GPT2-micro" GPT2 models on large Chinese corpus.}
\jhgreat{These sizes are set so the Transformer used in the Backpack's weight computation is the same size as the corresponding GPT2\hsr{-like Transformer model}.} 

\subsection{Data}

For pretraining, we employed three corpora: wiki2019zh~\cite{bright_xu_2019_3402023}, news2016zh~\cite{bright_xu_2019_3402023}, and webtext2019zh~\cite{bright_xu_2019_3402023}, which are composed of 1.04 million Wikipedia entries, 2.5 million news articles, and 4.1 million Q\&As, respectively, resulting in a total dataset size of 14.3G
\hsr{. This dataset} was used to pretrain ALBERT Chinese~\cite{albert_zh, lan2020albert}. To prepare the data, we set aside 1\% of the data for the test set and 0.5\% for the development set. The data was randomly partitioned into blocks of size 1,024 for each training step on each GPU.


\subsection{Evaluation method}
To evaluate the contextual performance of the Backpack and \hsr{Transformer} baseline models, we computed perplexities on 
\jhgreat{the test set of our web corpus.} We also 
\hsr{used} the Chinese WCPC dev set~\cite{ge-etal-2021-chinese}, \hsfixedsecond{an open-ended Chinese cloze task similar to LAMBADA~\cite{paperno-EtAl:2016:P16-1}, which includes 4,827 test cases and is 
\hsr{used} for assessing top-1 word accuracy in word prediction with long-term context}, to evaluate the models' ability to contextualize and predict words accurately. Specifically, each test case comprised a long sentence with at least 150 Chinese characters, with the last significant word being masked and having a length of 2 to 4 characters. The objective of the task was to predict the masked word, and we evaluated the performance of the models based on their top-1 and top-3 accuracy. As this task was originally tested on masked language models \hsfixedsecond{which 
\hsr{can see} 
the sentence's ending tokens, we designed a sampling method to evaluate our autoregressive models more fairly:} 
\hsfixedsecond{we generated characters with beam search until the length of the output tokens equaled the length of the original sentence. We retained ten generations from the beam in every step, penalized the outputs by adding the log probability of the original ending characters, and then selected the top generations.} 


\subsection{Experimental details}


The pretraining process for the Backpack-micro and the GPT2-micro baseline models involved training on 3$\times$ RTX3090 GPUs, using a batch size of 184,320 tokens for 500,000 gradient steps with cross-entropy loss, the AdamW~\cite{loshchilov2017decoupled} optimizer, 2,000 warmup steps, and linear decay on the learning rate starting from 6e-4 used by~\citet{nanoGPT}. The model with the best performance on the dev set was retained by evaluating at intervals of 1000 steps. The Transformer structure comprised 6 layers, 6 heads, and an embedding size of 384, with dropout disabled for flash attention~\cite{dao2022flashattention} in Torch 2.0. Three attempts were made to 
\hsr{improve} 
\hsr{parameterization} 
of the Backpack language model. 
\hsfixed{Compared to the original paper}, one layer was removed from the contextualization layer of the Transformer structure to match the size of the corresponding \hsr{Transformer} model. 
134M Backpack-small and GPT2-small were 
pretrained on \hsfixed{one A100 GPU} with a batch size of 245,760 tokens for 500,000 gradient steps, using 16 sense vectors and a Transformer structure comprising 12 layers, 12 heads, and an embedding size of 768.


\subsection{Results}

During the experiment, it was observed that pretraining the Backpack model was more challenging to stabilize compared to the 
\hsr{Transformer} model, although the overall loss curve of the 16-sense vector Backpack LM was similar to the 
\hsr{Transformer}. \hsfixedsecond{Specifically, in the Backpack architecture, the lack of layer normalization in the representation $\mathbf{o}_{i}$'s weighted sum computation can cause dramatic changes in the sense vectors and lead to gradient explosion during pretraining when encountering low-quality batches.} 

\hsr{In general, the Backpack models 
achieve similar perplexity scores compared to the GPT2-like Transformer model of similar scale and demonstrate significantly improved accuracy in WCPC}~\cite{ge-etal-2021-chinese} (Table~\ref{tab:model-performance}). 

WCPC is a challenging evaluation task as it requires the model to have long-distance contextualization ability and some world knowledge to determine the masked word. For the WCPC score, we found that our 134M Backpack-small tied with \hsfixed{223M} ALBERT-xxlarge 
Chinese~\cite{xu-etal-2020-clue} on top-1 accuracy and tied with the most performed MacBERT-large\cite{Cui_2021} in Chinese BERT family baselines~\cite{bert,liu2019roberta,cui-etal-2020-revisiting,Cui_2021} on top-3 accuracy using the ending words penalizing strategy. \hsfixed{Our strategy penalizes language models for generating predictions that do not end the sentence, improving evaluation alignment with masked language models.} 

\begin{table*}[ht]
\centering
\small
\begin{tabular}{ l c c c c }
\toprule
\textbf{Model} & \textbf{PPL} $\downarrow$ & \textbf{WCPC top-1 ACC} $\uparrow$ & \textbf{WCPC top-3 ACC} $\uparrow$ \\
\midrule
Backpack-micro & 16.25 & 2.98\% & 7.46\% \\ 
GPT2-micro & 16.66 & 2.44\% & 5.51\%  \\ 
\midrule
Backpack-small & 9.18 & 4.16\% & 10.6\% \\ 

GPT2-small  & 8.87 & 4.27\% & 10.42\% \\ 
\midrule
BERT-base, Chinese & - & 7.3\% & 10.1\% \\ 
RoBERTa-wwm-ext-base & - & 6.5\% & 9.8\% \\ 
MacBERT-large, Chinese & - & 6.8\% & 10.6\% \\ 
ALBERT-xxlarge, Chinese & - & 4.5\% & 6.5\% \\ 
\bottomrule
\end{tabular}
\caption{Language modeling performance. The baseline WCPC accuracies are from the original paper. For perplexity, lower is better; for accuracy, higher is better.}
\label{tab:model-performance}
\end{table*}

\section{Analysis of Lexical Structure}
\subsection{Sense Vectors}
\subsubsection{Visualizing Senses}
Following the Backpack paper, we projected the sense vectors of characters onto the vocabulary, denoted as $E^\top C(\mathbf{x})_{\ell} \in \mathbb{R}^{|\mathcal{V}|}$, to illustrate the contribution of the sense vectors towards predictions. \hsfixed{The outcomes are in Table \ref{tab:chars-short} and you can find a detailed version in the appendix (see Table \ref{tab:chars})}. As hypothesized, specific sense vectors \hsr{automatically} captured word composition rules \hsr{during pretraining}, whereas others captured semantic relatedness or associations.

\begin{table*}[ht]
\small
\begin{CJK}{UTF8}{gbsn}
\begin{minipage}{0.5\textwidth}
\centering
Sense Vector 10 (\textit{Word Composition}) \\
\begin{tabular}{c c}
\midrule
天 (sky / day) & 进 (enter / advance / come in) \\
\midrule
(天)涯 (distant land) & (进)驻 (settle in) \\
(天)津 (Tianjin City) & (进)入 (enter) \\
(天)竺 (Ancient India) & (进)军 (march) \\
(天)骄 (exceptional talent) & (进)攻 (attack) \\
(天)籁 (beautiful voice) & (进)展 (make progress) \\
\\
\end{tabular}
\end{minipage}%
\begin{minipage}{0.5\textwidth}
\centering
Sense Vector 12 (\textit{\hsr{Character} Meaning Relatedness}) \\
\begin{tabular}{c c}
\midrule
天 (sky / day) & 进 (\hsr{enter / advance / come in}) \\
\midrule
早 (early) & 步 (walk / step / pace)\\
夜 (night)& 必 (must / will / certainly)\\
醒 (wake up) & 毯 (blanket / carpet) \\
晚 (night) & 卧 (lie / crouch) \\
凌 (approach / rise high) & 洄 (eddy / whirlpool) \\
\\
\end{tabular}
\end{minipage}
\end{CJK}

\caption{The sense vectors in the same index learned a particular facet of character usage \hsr{in pretraining}. Each column contains the characters with the highest scores under the projection of the sense vectors on the vocabulary. \hsr{Sense vector 10 excels in composing two-character words, while sense vector 12 demonstrates strong character-level semantic correlations.}}
\label{tab:chars-short}
\end{table*}

\subsubsection{Word Representations}

In character-based languages, words are constructed through one or several characters in a complex manner. \hsfixedsecond{Linguistic studies have examined the morphological, orthographic, and phonological information within compound words~\cite{zhou1999morphology,packard2011new}. However, we distinguish them into the following categories based on whether the characters convey meaning individually and the implicit information density within the characters.} In detail, some words are composed of characters with sub-meanings \hsfixedsecond{("compound word")}, while some borrowed words from foreign languages only use the pronunciations of the characters ("loanword"). There are also four-character words that represent lengthy allusions, with the characters representing the critical objects in the allusion ("idiom"). 

\hsfixed{
We explored methods for better representing these vocabularies based on the sense vectors of the compositional characters to test lexical relationship on the words with explicit meanings. Here are the two methods that we explored. 

Firstly, We purposed a method which involved simply computing the average value of the sense vectors of the constituent characters to represent the word's sense vector.

Secondly, we hypothesize that 
\hsr{words with a complicated, non-systematic function from characters to the word meaning} will have their constituent character senses weighted similarly no matter what context they appear in---thus constructing the non-systematic meaning. \hsfixedsecond{
Suppose we have \hsr{a} context ${\mathbf{c}}$ that \hsr{contains} a target word with p constituent characters $w = \x_1,\dots,\x_p$, with the index of these characters in the context ${\mathbf{c}}$ as $j_{\x_1},\dots,j_{\x_p}$, we calculate the average contextual composition ratio $\lambda(\mathbf{c})_{\ell}$ \hsr{on sense vector $\ell$ as
\begin{align}
    \frac{\lambda(\mathbf{c})_{\ell j_{\x_1}}}{\sum_{s=1}^{p} \lambda(\mathbf{c})_{\ell j_{\x_s}}}, \dots, \frac{\lambda(\mathbf{c})_{\ell j_{\x_p}}}{\sum_{s=1}^{p} \lambda(\mathbf{c})_{\ell j_{\x_s}}}
\end{align}
where}
\hsr{\begin{align}
    \lambda(\mathbf{c})_{\ell j_{\x_s}} = \frac{1}{|\mathbf{c}| - j_{\x_p}} \sum_{i = j_{\x_p} + 1}^{|\mathbf{c}| + 1} \frac{\alpha_{\ell ij_{\x_s}}}{\sum_{k=1}^{p} \alpha_{\ell ij_{\x_k}}}
\end{align}}
We expect the ratios $\lambda(\mathbf{c}_1)_{\ell} \approx \dots \approx \lambda(\mathbf{c}_q)_{\ell}$ for any q contexts without any significant semantic amplifications on the meaning any of the constituent characters.
Assuming this hypothesis holds, a word $w$ could be represented as 
\hsr{\begin{align}
    C(w)_\ell = \frac{1}{q} \sum_{m=1}^{q} \sum_{s=1}^{p} \lambda(\mathbf{c}_m)_{\ell j_{\x_s}} C(\mathbf{x}_s)_\ell 
\end{align}}
for samples of context $\mathbf{c}_1, \dots, \mathbf{c}_q$.
}
}

To prove the feasibility of the second method, we designed several prompts (Appendix \ref{tab:prompts}) that fit different types of words and calculate the average contribution ratio of each character's sense vectors among all constituent characters in the word and how much each contribution is away from the average value. \hsfixedsecond{We created a dataset containing 120 compound words, 102 loanwords, and 104 idioms, and validated the above hypothesis on this dataset.} Our experimental results showed that each character's contribution ratio in a word on each sense vector for prediction remained stable across various contexts. Furthermore, the stability of word compositions was observed to follow the order of \hsfixedsecond{idiom $>$ compound word $>$ loanword as shown in Table~\ref{tab:contribution}. However, we also observed that while the senses of most vocabulary items are highly stable across different contexts, there exists a subset of vocabulary items that exhibit poor stabilities. The underlying reasons for this phenomenon warrant further investigation. More word examples are in the Appendix~\ref{tab:contribution_word}.} 

\begin{table}[t]
\small
\centering
\begin{tabular}{c c c c}
\toprule
\textbf{Type} & \textbf{$\leq \pm10\%$} & \textbf{$\leq \pm20\%$} & \textbf{$\geq \pm20\%$} \\ \midrule
compound words & 69.53\% & 26.61\% & 4.06\% \\ \midrule
loanwords & 60.60\% & 29.64\% & 9.76\% \\ \midrule
idioms & 84.94\% & 13.94\% & 1.20\% \\ \bottomrule
\end{tabular}
\caption{How the contribution ratio of sense vectors on characters of a word varies among the different contexts. A more minor variation in the contribution ratio indicates a more stable word composition.}
\label{tab:contribution}
\end{table}

\subsubsection{Lexical Relationship Test}

We evaluated the lexical relationship of the sense vectors using two datasets: Wordsim-240 and Wordsim-297 \cite{sememe}, 
and represent a word by averaging all the sense vectors of the constituent characters. 
To assess the quality of the resulting lexical representations, we computed Spearman rank-order correlation coefficient 
between the relationship scores in the datasets and the cosine similarities of each word pair across all the sense vectors of our models. For the GPT2 model, we represented each word by averaging the embeddings of the constituent characters.

Our results in table \ref{tab:relation} show that our Backpack Model outperformed the same-scaled GPT2 model, but the results were significantly inferior to word embeddings trained \textit{directly on words} using methods such as word2vec \cite{word2vec} or GLoVe \cite{pennington2014glove}.

\begin{table}[ht]
\centering
\small
\begin{tabular}{l c c c c}
\toprule
\textbf{Representation} & \textbf{WS240} & \textbf{WS297} \\ \midrule
\cjktext{Backpack-micro \#14} & 0.335 & 0.226 \\ \midrule
\cjktext{GPT2-micro} & 0.164 & 0.271 \\ \midrule
\cjktext{Backpack-small \#9} & \textbf{0.384} & \textbf{0.426} \\ \midrule
\cjktext{GPT2-small} & 0.225 & 0.334 \\ \midrule
\textit{Word-tokenized models} & & \\
\textit{(not comparable)} & & \\
\cjktext{CBOW} & 0.561 & 0.626 \\ 
\cjktext{GloVe} & 0.558 & 0.584 \\ \bottomrule
\end{tabular}
\caption{Pearson product-moment correlation coefficients between the provided scores and the cosine similarities of the word pairs are calculated. Character-tokenized Backpack LMs outperform GPT2 but are inferior to word-tokenized models.
}
\label{tab:relation}
\end{table}


\begin{table*}[ht]
\begin{CJK}{UTF8}{gbsn}
\centering
\small
\begin{tabular}{ c c c c c c c }
\toprule
 & & \multicolumn{2}{c}{\bf Transformers} & \multicolumn{3}{c}{\bf Backpacks (ours)}\\
 \cmidrule{3-4} \cmidrule{5-7}
\textbf{Character} & \textbf{Target Word} & \textbf{GPT2} & \textbf{GPT2 proj} & \textbf{Backpack} & \textbf{half \#15} & \textbf{remove \#15} \\ \midrule
\cjktext{兵} (arms) & \cjktext{士兵} (soldier) & 70.32 & 55.55 & 58.13 & 34.95 & 21.34 \\ \midrule
\cjktext{警} (alert) & \cjktext{警察} (police) & 20.93 & 20.47 & 23.62 & 14.90 & 9.47 \\ \midrule
\cjktext{演} (act) & \cjktext{演员} (actor / actress) & 6.58 & 6.19 & 4.92 & 4.50 & 4.13 \\ \midrule
\cjktext{会} (teach) & \cjktext{教师} (teacher) & 2.45 & 2.40 & 4.69 & 4.13 & 3.65 \\ \midrule
\end{tabular}
\caption{Character-level bias ratio; by partially or totally removing sense 15, the character and the words composed by the character get debiased. A perfect unbiased model would achieve a ratio of 1.}
\label{tab:debias}
\end{CJK}
\end{table*}

\begin{table*}[ht]
\small
\begin{CJK}{UTF8}{gbsn}
\centering
\begin{tabular}{c c c c c c c c c c}
\toprule
\textbf{Multipliers} & \cjktext{撒} & \cjktext{粒} & \cjktext{堡} & \cjktext{丘} & \cjktext{石} & \cjktext{人} & \cjktext{球} & \cjktext{海} & \cjktext{晒} \\
沙(sand),滩(beach) & (\hsr{sanding}) & (particle) & (castle) & (dune) & (stone) & (people) & (ball) & (sea) & (bask) \\ \midrule
1,1 & 1 & 1 & 1 & 1 & 1 & 1 & 1 & 1 & 1 \\ \midrule
4,1 & 2.13 & 1.74 & 1.42 & 1.27 & 1.14 & 0.78 & 0.71 & 0.62 & 0.61 \\ \midrule
1,4 & 0.54 & 0.55 & 0.70 & 0.71 & 0.71 & 1.23 & 1.25 & 1.24 & 1.48 \\ 
\bottomrule
\end{tabular}
\caption{\hsr{The ratio of probabilities on predicting certain characters by amplifying the sense vectors with multipliers for the characters "沙" (sand) and "滩" (beach) compared to the original probabilities.}}

\label{tab:centroidEval}
\end{CJK}
\end{table*}

\subsection{Sense Vectors for Control}
In this section, we showcase two character-level interventions on the sense vectors as proof-of-concept. 

\subsubsection{Mitigating gender bias}
In Modern Chinese, most professions are composed of two or more Chinese characters, making direct debiasing of stereotypically gendered profession nouns difficult. To address this issue, we attempted two approaches: 1) identifying the characters within the composed words that contain gender-biased meanings and debiasing them from their sense vectors, and 2) directly debiasing the sense vectors of the composed words using the method discussed in Word Representations.

\begin{CJK}{UTF8}{gbsn}
We hypothesized that the first approach could be practical because many Modern Chinese words are combined from ancient single-character words that represent a relevant meaning to the composed words. For example, the word "\cjktext{士兵}" (soldier) is composed of "\cjktext{士}" (man/warrior) and "\cjktext{兵}" (arms), both of which carry stereotypical male bias. In our experiments, we attempted to identify the sense vectors of characters that contain gender stereotypes and compared \hsr{$\left |(EC(\mathbf{x_{he}})_{\ell} - EC(\mathbf{x_{she}})_{\ell})\right|$} 
to determine which sense vectors contribute to gender bias. We found that sense 3 contributed the most bias. Using the method described in the Backpack paper, we reduced the weight of sense 3 on these characters. We evaluated how the composed words were gender debiased by creating several prompts (Appendix \ref{tab:prompts2}) that fit all the profession words, filling in the target word, and computing the average bias probability score of "\cjktext{他} (he/him)" versus "\cjktext{她} (she/her)" as $\mathbb{E}_{X \in \text{prompts}}[\text{max} (\frac{p(\text{he} | \mathbf{x})}{p(\text{she} | \mathbf{x})}, \frac{p(\text{she} | \mathbf{x})}{p(\text{he} | \mathbf{x})})]$.

\paragraph{Baseline.} 
We employed a similar approach as described in the Backpack paper, which was inspired by the work of \cite{Bolukbasi2016ManIT}. 
Specifically, we computed the gender bias direction using the difference between the embeddings of the words "\cjktext{他} (he/him)" and "\cjktext{她} (she/her)," denoted as $E\mathbf{X_{he}} - E\mathbf{X_{she}}$, and then projected the embeddings of the biased characters onto the nullspace of this direction.
\end{CJK}

\paragraph{Results.} 
We experimented with investigating the effect of removing sense 15 from several characters on bias scores of profession words containing those characters. The bias ratios resulting from this experiment are reported in the table~\ref{tab:debias}. Our experimentation demonstrated that removing sense 15 substantially decreased the bias in words that were originally more biased while producing a considerably lesser impact on words with lower levels of bias. Nonetheless, this approach yielded significant improvements compared to the GPT2 baseline.

Besides, we explored the second approach by removing sense 15 for both constituent characters. Surprisingly, this approach was less effective than the first approach. To investigate whether there exists a specific sense vector to remove for all characters in all compositional words for gender debiasing, we experimented and observed that reducing sense 3 significantly reduced the bias in the word \cjktext{警察} (police); however, the reducing sense 3 method did not generalize to other words. We hypothesize that the model might not effectively learn the gender-representing information due to the limited model size and pretraining steps. Some critical gender-related information might still distribute among several sense vectors.


\subsubsection{Character Amplification Control}

Focusing on sub-meanings or properties in a word constructed by multiple characters makes more sense in character-based languages. For instance, the Chinese word "\cjktext{词典}" which means "dictionary," is composed of the characters "\cjktext{词}" (word) and "\cjktext{典}" (book, in ancient Chinese), and when generating text from input containing this word, the model could focus on either the "word" or "book" property. By adjusting the weights of the sense vectors of the constituent characters, we were able to amplify implicit meaning of a constituent character and bias the model toward generating text related to a specific property. Specifically, we conducted experiments to amplify the contribution of the first or second character four times each while keeping the total contribution of the word unchanged in the output. We found that the model tended to generate sentences that relate to the amplified character with greater probability, as shown in Appendix \ref{tab:centroid}. We assessed the efficacy of the proposed method by computing the ratio of expectations for the controlled model relative to an uncontrolled model in the context of predicting semantically related characters from an open-topic prompt as $\mathbb{E}_{\mathbf{c}_{target}}[\frac{p(\mathbf{c}_{target} | \mathbf{x}_{amp})}{p(\mathbf{c}_{target} | \mathbf{x})}]$. Table \ref{tab:centroidEval} illustrates an instance of the outcome of amplifying characters in the word "\cjktext{沙滩}" (beach). Notably, the findings indicate that character-specific semantics were the most amplified. We hypothesize that this work can assist in scenarios where it is necessary to precisely generate expressions that convey the author's intended meaning in a short sequence, such as poetry, songwriting, or beginning a discourse around one of the meanings in a polysemous word.


\section{Conclusion}
In this paper, we presented implementing, pretraining, and evaluating a character-based Chinese Backpack language model. We conducted extensive experiments on sense vector visualizations, word representations, lexical relationships, and idiom compositions and explored two approaches to character-level interventions. Our results demonstrate the potential of Backpack LM in language modeling tasks for character-based languages, the interpretability of the sense vectors on the character and word level, and the potential of character-level interventions across various contexts.

\section{Limitations}

Despite these promising results, there are several limitations to our study. First, we had limited GPU resources, which prevented us from attempting a larger batch size during pretraining. Second, our word interventions depend on the sub-meanings of the characters, and we currently have no solution to effectively intervene in transliterated words by modifying the sense vectors of the characters that only represent phonetic information. Therefore, intervening in character-based languages where many words are transliterated, such as Korean, remains challenging. Third, we observed that although our approach enables greater flexibility in character-level sense vectors to represent richer morphological structures, word representations by characters are less interpretable than word sense vectors learned by models using word tokenizations, particularly for complex words such as idiomatic phrases. We believe that this issue could be mitigated by increasing the number of sense vectors with a larger contextualization model and pretraining with more data. Further research is required to address these limitations and explore the potential of word representations and interventions in character-based languages.

\bibliography{anthology,custom}
\bibliographystyle{acl_natbib}

\appendix
\section{Language Modeling Details}\label{sec:modeling}
\subsection{Residual Connection}
We started our experiment with no second residual connection. \hsfixed{However, we found that adding second residual connection by unsqueezing the output from the first feed-forward layer by dimension $k$ to match $k * d$ dimensions improved training stability compared to the specification of \citet{backpack}.}

\subsection{Comparison of Parameter Numbers}
The contextualization weight function was defined with mask filling and an extra dropout layer included after the Softmax function. 

\hsfixedsecond{To make a fair comparison with the corresponding GPT2 model, we analysed the number of parameters and removed one block from the Transformer structure of the Backpack model. As discussed, the contextualization weight of each sense vector is calculated with additional matrices $K, Q \in \mathbb{R}^{d \times d}$. The first feed forward layer in the sense vector layer involves an up projection matrix $\in \mathbb{R}^{d \times 4d}$ and a down projection matrix $\in \mathbb{R}^{4d \times d}$. Summing up these parameters, we have a ${10d^2}$ additional parameter size, which is close to the ${12d^2}$ parameter size in a single Transformer block so that by removing one block, we will only add $(k - 2) * d^2 \approx k * d^2$ parameters which are necessary for representing the sense vectors.}

\section{Interpreting Idiom Composition}
\begin{CJK}{UTF8}{gbsn}
We investigated which sense vectors played a dominant role when the model used the first three characters of idiomatic phrases as input to predict the last character. However, we encountered difficulty in interpreting the character composition of idiomatic phrases. For example, when analyzing the phrase "\cjktext{画蛇添(足)}" i.e., "drawing legs on a snake," which means "an unnecessary and redundant act that spoils the original effect or even makes it worse," by stacking weights of the first three characters on 16 or 64 sense vectors, we found that using any single sense vector for prediction did not significantly lead the model to output the target character, even though the model correctly outputted "\cjktext{足}" i.e., "leg" after performing a weighted sum of these sense vectors. We projected 16 sense vectors onto the vocabulary and examined their projections onto the character; however, we observed that none exhibited a disproportionately large or small projection onto the resulting character. \hsfixed{This experiment provides evidence that the top components of sense vectors may not effectively capture how they will compose to make predictions.}
\end{CJK}

\begin{table*}[ht]
\small
\centering
\begin{CJK}{UTF8}{gbsn}
Sense Vector 10 (\textit{Word Composition}) \\
\begin{tabular}{c c c c}
\midrule
天 (sky / day) & 沙 (sand) & 进 (enter / advance / come in) & 自 (from / self) \\
\midrule
(天)涯 (distant land) & (沙)漠 (desert) & (进)驻 (settle in) & (自)由 (freedom) \\
(天)津 (Tianjin City) & (沙)鸥 (gull) & (进)入 (enter) & (自)慰 (console) \\
(天)竺 (Ancient India) & (沙)哑 (hoarse) & (进)军 (march) & (自)如 (the App Ziroom) \\
(天)骄 (exceptional talent) & (沙)溢 (actor Yi Sha) & (进)攻 (attack) & (自)拍 (selfie) \\
(天)籁 (beautiful voice) & (沙)滩 (Beach) & (进)展 (make progress) & (自)卸 (self-dumping) \\
\\
\end{tabular}

Sense Vector 12 (\textit{\hsr{Character} Meaning Relatedness or Composition}) \\
\begin{tabular}{c c c c}
\midrule
天 (sky / day) & 沙 (sand) & 进 (enter / advance / come in) & 自 (from / self) \\
\midrule
早 (early) & 04 (FC Schalke 04) & 步 (walk / step / pace) & 从 (from)\\
夜 (night) & (沙)箱 (sandbox) & 必 (must / will / certainly) & 之 (he / she / it / go / 's)\\
醒 (wake up) & (沙)盒 (sandbox) & 毯 (blanket / carpet) & 打 (since)\\
晚 (night) & 毒 (poison) & 卧 (lie / crouch) & 感 (sense / feel)\\
凌 (approach / rise high) & 铂 (platinum) & 洄 (eddy / whirlpool) & 蚂 (ant)\\
\\
\end{tabular}

Sense Vector 15 (\textit{\hsr{Character} Meaning Relatedness or Composition}) \\
\begin{tabular}{c c c c}
\midrule
天 (sky / day) & 沙 (sand) & 进 (enter / advance / come in) & 自 (from / self) \\
\midrule
黑 (black) & 潇 (drizzle) & (进)展 (progress) & (自)大 (arrogant)\\
亮 (light) & 浏 (clear) & 顺 (smooth) & (自)满 (complacent)\\
昨 (yesterday) & 湖 (lake) & 神 (magical / god) & 狠 (ruthless)\\
黑 (black) & 岳 (mountain) & 慢 (slow) & (自)暴 (Give up on yourself)\\
今 (today) & 橘 (tangerine) & 缓 (delay) & (自)免 (to resign voluntarily)\\
\\
\end{tabular}
\end{CJK}

\caption{The sense vectors in the same index are considered to have a particular facet of character usage. Each column contains the characters with the highest scores under the projection of the sense vectors on the vocabulary.}
\label{tab:chars}
\end{table*}

\begin{table*}[ht]
\small
\centering
\begin{tabular}{c c c c c c}
\toprule
\textbf{Type} & \textbf{Word} & \textbf{Stablility} & \textbf{$\leq \pm10\%$} & \textbf{$\leq \pm20\%$} & \textbf{$\geq \pm20\%$} \\ \midrule
compound& \cjktext{手机} (telephone) $=$ \cjktext{手} (hand) + \cjktext{机} (machine) & high & 16 & 0 & 0 \\ 
words & \cjktext{大学} (university) $=$ \cjktext{大} (large) + \cjktext{学} (learn) & high  & 16 & 0 &0 \\ 
& \cjktext{孤独} (lonely) $=$ \cjktext{孤} (isolated) + \cjktext{独} (alone) & low  & 1 & 6 & 9 \\ 
\midrule

& \cjktext{马赛克} (Mosaic) & high & 16 & 0 & 0 \\ 
loanwords & \cjktext{迷你} (mini) & high  & 12 & 4 &0 \\ 
& \cjktext{夸克} (quark) & low  & 5 & 7 & 4 \\ 
\midrule

& \cjktext{骑虎难下} (in a difficult situation with no easy way out) & high & 16 & 0 & 0 \\ 
idioms & \cjktext{画蛇添足}(to do something unnecessary even harmful) & high  & 14 & 2 &0 \\ 
& \cjktext{韬光养晦} (to wait for the right moment to shine) & low  & 12 & 2 & 2
\\ \bottomrule
\end{tabular}
\caption{How many sense vectors for each range of the contribution ratio on characters of a word varies among the different contexts. A more minor variation in the contribution ratio indicates a more stable word composition.}
\label{tab:contribution_word}
\end{table*}

\begin{table*}[ht]
\begin{CJK}{UTF8}{gbsn}
\centering
\small
\begin{tabular}{c c}
\midrule
\textbf{prompt} & \textbf{English} \\ \midrule
\cjktext{WORD} & WORD \\
\cjktext{"WORD"的意思是} & The meaning of "WORD" is \\ 
\cjktext{老师曾教育，WORD} & A teacher told that WORD \\ 
\cjktext{关于WORD，} & About WORD, \\ 
\cjktext{电视里说，WORD} & In TV, it is said that WORD \\
\cjktext{WORD是} & WORD is \\
\cjktext{我觉得WORD} & I think WORD \\ \midrule
\end{tabular}
\caption{General prompts for different type of nouns}
\label{tab:prompts}
\end{CJK}
\end{table*}

\begin{table*}[ht]
\begin{CJK}{UTF8}{gbsn}
\centering
\small
\begin{tabular}{c c}
\midrule
\textbf{prompt} & \textbf{English} \\ \midrule
\cjktext{那个WORD说，} & That WORD said, \\ 
\cjktext{这个WORD相信} & This WORD believes \\ 
\cjktext{WORD进到屋子里，} & The WORD enters the house, \\ 
\cjktext{WORD坐在车里，然后} & The WORD sat in the car, and then \\
\cjktext{WORD走了过来，} & Then WORD came over,\\ \midrule
\end{tabular}
\caption{General prompts for gender bias evaluations}
\label{tab:prompts2}
\end{CJK}
\end{table*}

\begin{table*}[ht]
\begin{CJK}{UTF8}{gbsn}
\centering
\small
\begin{tabular}{c c c}
\midrule
\textbf{Word} & \textbf{Multiplier} & \textbf{Output} \\ \midrule
\cjktext{沙滩} (beach) & 1,1 & \cjktext{沙滩上有很多人。} \\ \cjktext{沙(sand) 滩(beach / puddle}) & & (There are a lot of people on the beach.) \\ 
\cjktext{沙滩} (beach) & 4,1 & \cjktext{沙滩上有很多大大小小的沙堆。} \\ \cjktext{沙(\textbf{sand}) 滩(beach / puddle}) & & (On the beach, there are many big and small \textbf{sand dunes}.) \\ 
\cjktext{沙滩} (beach) & 1,4 & \cjktext{沙滩上有很多人在海边钓鱼。} \\ \cjktext{沙(sand) 滩(\textbf{beach / puddle}}) & & (There are many people fishing \textbf{by the seaside} on the beach.) \\ \midrule
\cjktext{理想} (ideal) & 1,1 & \cjktext{理想是什么?我很迷茫，不知道自己喜欢什么。} \\ \cjktext{理(principle / logic) 想(imagine / want}) & & (What is ideal? I am confused and unsure of what I truly like.) \\ 
\cjktext{理想} (ideal) & 4,1 & \cjktext{理想是什么？如何理解？} \\ \cjktext{理(\textbf{principle / logic}) 想(imagine / want}) & & (What is ideal? How to \textbf{understand} it?) \\ 
\cjktext{理想} (ideal) & 1,4 & \cjktext{理想是什么？如何做到？} \\ \cjktext{理(principle / logic) 想(\textbf{imagine / want}}) & & (What is ideal? How to \textbf{achieve} it?) \\ \midrule
\end{tabular}
\caption{Generative outputs on the character amplification control task with top probabilities. Note that the word "理想" means "ideal" but is combined with the characters meaning "principle / logic" and "imagine / want".}
\label{tab:centroid}
\end{CJK}
\end{table*}
\end{CJK}
\end{document}